\documentclass{article}

\usepackage{iclr2027_conference,times}

\usepackage{amsmath,amsfonts,bm}

\def\eqref#1{equation~\ref{#1}}

\def\1{\bm{1}}

\DeclareMathAlphabet{\mathsfit}{\encodingdefault}{\sfdefault}{m}{sl}
\SetMathAlphabet{\mathsfit}{bold}{\encodingdefault}{\sfdefault}{bx}{n}

\usepackage{hyperref}
\hypersetup{hidelinks}
\usepackage{url}
\usepackage{booktabs}
\usepackage{tabularx}
\usepackage{longtable}
\usepackage{needspace}
\usepackage{graphicx}
\usepackage{float}
\usepackage{caption}
\usepackage{xcolor}
\usepackage{colortbl}
\usepackage{placeins}
\usepackage{amssymb}
\newcommand{\ours}{FutureDuet}
\definecolor{bestcell}{RGB}{226,239,204}
\definecolor{secondcell}{RGB}{250,244,216}
\newcommand{\best}[1]{\cellcolor{bestcell}\textbf{#1}}
\newcommand{\second}[1]{\cellcolor{secondcell}\underline{#1}}

\title{FutureDuet: Decoupling Observation\\Access from Future Supervision in\\ World Action Models}

\author{
Jie Wu$^{1,2}$\thanks{Work done during an internship at TeleAI.}\quad
Yuzhi Huang$^{1}$\quad
Junqi Liu$^{2}$\quad
Weichen Zhang$^{1}$\\
Haibin Huang$^{2}$\quad
Yin Chen$^{2}$\quad
Jingyan Jiang$^{3,\dagger}$\quad
Chi Zhang$^{2,\dagger}$\\
$^{1}$Tsinghua University\quad
$^{2}$TeleAI, China Telecom\quad
$^{3}$Shenzhen Technology University\\
$^{\dagger}$Corresponding authors.\\
\textit{Project website:} \url{https://1723578110.github.io/futureduet-web/}
}

\iclrfinalcopy

\begin{document}
\maketitle
\lhead{Preprint}

\begin{abstract}
World Action Models (WAMs) augment robot action generation with future visual supervision. Existing WAMs commonly fuse main and wrist observations into one visual stream and train both with the same future-video objective, despite their different visual dynamics. A stable main camera reveals scene-level task evolution, whereas wrist cameras move with the end effector, mixing local interaction changes with viewpoint shifts and self-occlusion. These contrasting predictive demands suggest that the two views may benefit from different future objectives. We introduce \ours, which retains both views for control, while allowing each visual stream to receive a different future objective. For the main view, future RGB models task evolution, while interaction masks and robot skeletons focus supervision on task objects and robot motion. For the wrist stream, future latent prediction models short-horizon interaction changes without requiring pixel-level reconstruction. ActionDiT jointly reads the resulting Task State and Interaction State, combining scene-level progress with close-range interaction evidence. All auxiliary prediction modules are training-only, adding no inference overhead. \ours{} achieves 94.2\% clean and 94.1\% randomized success on RoboTwin50 and 99.2\% average success on LIBERO. The improvements are most pronounced on six RoboTwin50 tasks that require precise interaction, averaging gains of 9.2\% and 12.8\% over Fast-WAM in clean and randomized settings. Controlled studies further show complementary gains from separating the wrist pathway and designing future supervision separately for the two views.
\end{abstract}

\section{Introduction}
\label{sec:introduction}

World Action Models (WAMs) augment robot action generation with future visual supervision, leveraging predicted scene evolution to improve policy learning~\citep{bi2025motus,yuan2026fast,cen2025worldvla,kim2026cosmos,ma2026dit4dit,cheang2024gr2generativevideolanguageactionmodel,cen2025rynnvla,dreamvla25,pai2025mimic}. Fast-WAM further shows that this benefit remains largely when future observations are not generated at deployment~\citep{yuan2026fast}. This finding shifts attention from the generation of futures at test time to how future supervision should be designed during training.\footnote{For broader system-level perspectives on integrating AI with sensing, communication, and control, see the AI Flow framework~\citep{aiflowPerspective,aiflowEdge}.}

\begin{figure}[t]
    \centering
    \includegraphics[width=\linewidth]{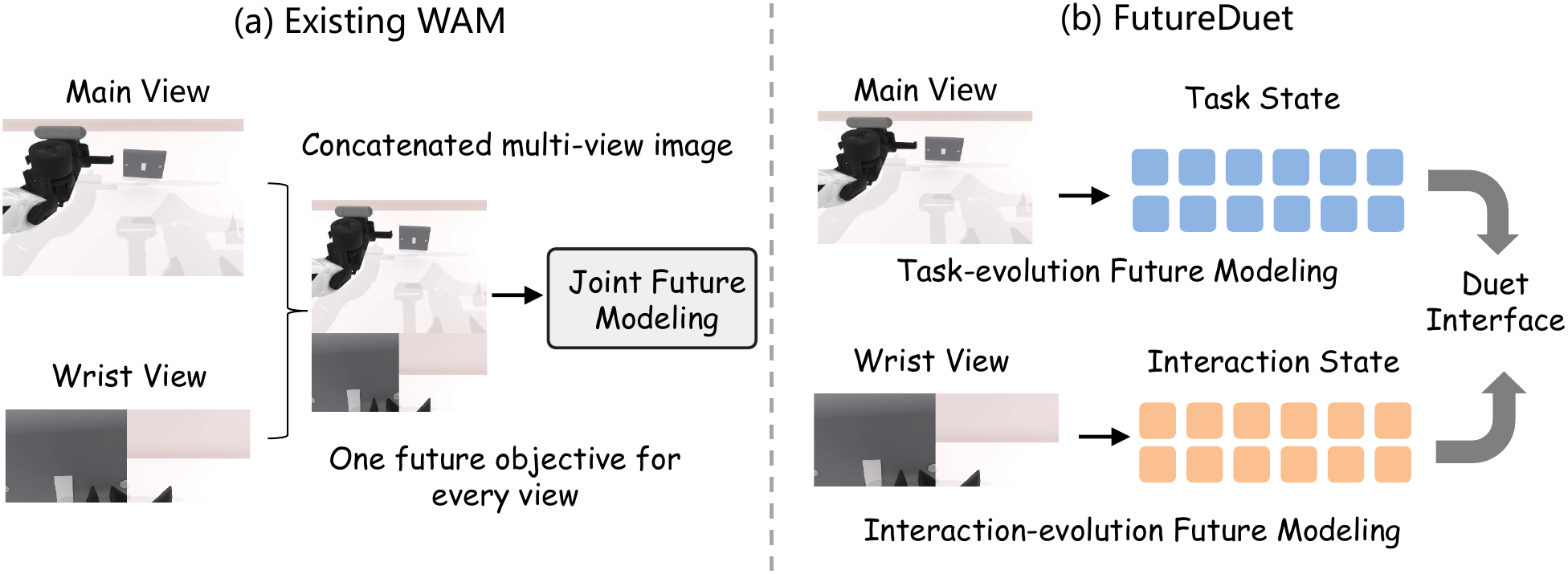}
    \caption{\textbf{Decoupling future supervision across views.} (a) A common multi-view WAM design concatenates observations and applies one future objective across views. (b) \ours\ separately models task evolution from the main view and interaction evolution from wrist views; the resulting states jointly support action generation.}
    \label{fig:view-modeling}
\end{figure}

For WAMs that use both main and wrist cameras, a common design is to fuse their observations into a shared visual stream and train both views under a uniform future-video objective, as illustrated in Fig.~\ref{fig:view-modeling}(a)~\citep{bi2025motus,yuan2026fast,cen2025worldvla}. This makes incorporating both views straightforward, but entangles observation access with future supervision: every camera stream available to the policy is forced to predict the same form of future target.

However, the camera’s frame of reference changes the prediction problem itself. A main camera provides a relatively stable scene frame in which object relations, robot motion, and task progress can be tracked over time. Wrist cameras instead move with the end effector, bringing relative pose, contact, and object response into close view while also introducing rapid viewpoint shifts and self-occlusion. From the main view, future changes reveal how the task unfolds across the scene; in wrist views, they reflect both the local interaction and the motion of the camera itself. The two views therefore support complementary state representations for control: one tracks scene-level task progress, while the other captures the ongoing close-range interaction. A shared future video objective nevertheless treats these different visual changes as the same form of prediction target, raising a central question: should each view model the same kind of future?

We therefore introduce \ours{} (Fig.~\ref{fig:view-modeling}(b)), a WAM that keeps both main and wrist observations available for action generation while allowing their future supervision to be designed separately. For the main view, future RGB prediction captures how the task evolves across the scene, while interaction masks and robot skeletons focus the representation on task objects and robot motion. For the wrist views, future latent prediction models short-horizon interaction changes without requiring pixel-level reconstruction of appearance changes caused by camera motion and self-occlusion. The two streams consequently provide complementary control information: the \emph{Task State} tracks scene-level task progress, while the \emph{Interaction State} captures close-range interaction changes. ActionDiT reads both states jointly to generate actions. All auxiliary prediction modules operate exclusively during training, adding no inference overhead.

\ours{} achieves 94.2\% clean and 94.1\% randomized success on RoboTwin50 and 99.2\% average success on LIBERO, with its largest RoboTwin50 gains on tasks requiring precise interaction completion. Controlled ablations show gains from wrist-pathway separation; with the deployed architecture fixed, future supervision provides further gains and latent targets perform best. Structured main-view supervision provides complementary gains, while readout diagnostics connect masks and skeletons to object- and robot-related evidence.

Our contributions are:
\begin{itemize}
    \item We propose \ours, a WAM that decouples observation access from future supervision. It retains main and wrist observations for action generation while allowing their future objectives to be designed separately. The auxiliary prediction modules operate only during training, adding no inference overhead.

    \item We design two complementary streams for joint control. The main stream learns a \emph{Task State} through future RGB prediction, with masks and skeletons emphasizing task objects and robot motion. The wrist stream learns an \emph{Interaction State} through future latent prediction to capture close-range interaction changes.

    \item We validate \ours{} on RoboTwin50 and LIBERO, with the largest RoboTwin50 gains on tasks requiring precise interaction completion. Our controlled wrist-pathway study shows a 13.9-percentage-point improvement over unified RGB modeling, highlighting the benefits of separate wrist modeling and future supervision.
\end{itemize}
\section{Related Work}
\label{sec:related-work}

\noindent\textbf{Vision-Language-Action Policies and World Action Models.}
Vision-Language-Action policies map visual observations and language instructions to robot actions, drawing on pretrained vision-language representations and diverse robot demonstrations~\citep{brohan2023rt2,octoteam2024octo,kim2024openvla,black2024pi0,bjorck2025gr00t,qu2025spatialvla,liu2025rdt,song2025reconvlareconstructivevisionlanguageactionmodel,wu2026pragmatic}. World Action Models augment this mapping with visual future modeling, allowing future observations or representations to shape action learning~\citep{du2023learninguniversalpoliciestextguided,zhou2024robodreamer,bharadhwaj2024gen2act,feng2025vidarembodiedvideodiffusion,jang2025dreamgenunlockinggeneralizationrobot}. Existing methods use this predictive component in different ways. Some generate future observations as intermediate plans for action prediction~\citep{du2023learninguniversalpoliciestextguided,zhou2024robodreamer,bharadhwaj2024gen2act,liang2025videogenerators,wu2023unleashing}, while others jointly model video and action trajectories~\citep{bi2025motus,cen2025worldvla,zhu2025unifiedworldmodelscoupling,won2025dualstreamdiffusionworldmodelaugmented}. Fast-WAM further shows that future prediction can serve as an effective training signal without generating future observations during deployment~\citep{yuan2026fast}. Future modeling in WAMs therefore serves not only as an inference mechanism, but also as a source of supervision for learning robot policies.

\noindent\textbf{Predictive Targets for Robot Control.}
The choice of future target determines which aspects of change a policy is trained to model. RGB targets describe future scene appearance and motion~\citep{hu2024video,yuan2026fast}, whereas latent and joint-embedding objectives model future visual representations without reconstructing every pixel~\citep{sun2026vla,zhang2026disentangled}. Recent WAMs use compact latent futures or pretrained joint-embedding spaces to capture predictive dynamics~\citep{chen2026lawam,huang2026forewam,lin2026jepawam}. Concurrent work W2-VLA predicts task-conditioned future wrist latents and exposes the resulting future-aware context to action generation~\citep{pan2026worldtowrist}. Other approaches emphasize action-relevant structure through object masks, point trajectories, or complementary semantic and geometric targets~\citep{maskwam,bharadhwaj2024track2act,haldar2025point,yuan2026dreamwam}. \ours\ uses wrist-latent prediction only as training supervision and examines a broader design question: how different future objectives can be assigned across main and wrist streams that are jointly used for action generation.

\section{Method}
\label{sec:method}

\subsection{Overview}
At control step $t$, \ours\ accepts a language instruction $c$, proprioception $p_t$, main-view history $o^m_{\leq t}$, and wrist histories $o^w_{\leq t}$ to generate an action chunk $a=a_{t:t+H-1}$. Both views support control, but their future objectives are designed separately: the main stream models scene-level task evolution, while the wrist stream models short-horizon interaction changes.

\begin{figure}[t]
    \centering
    \includegraphics[width=\linewidth]{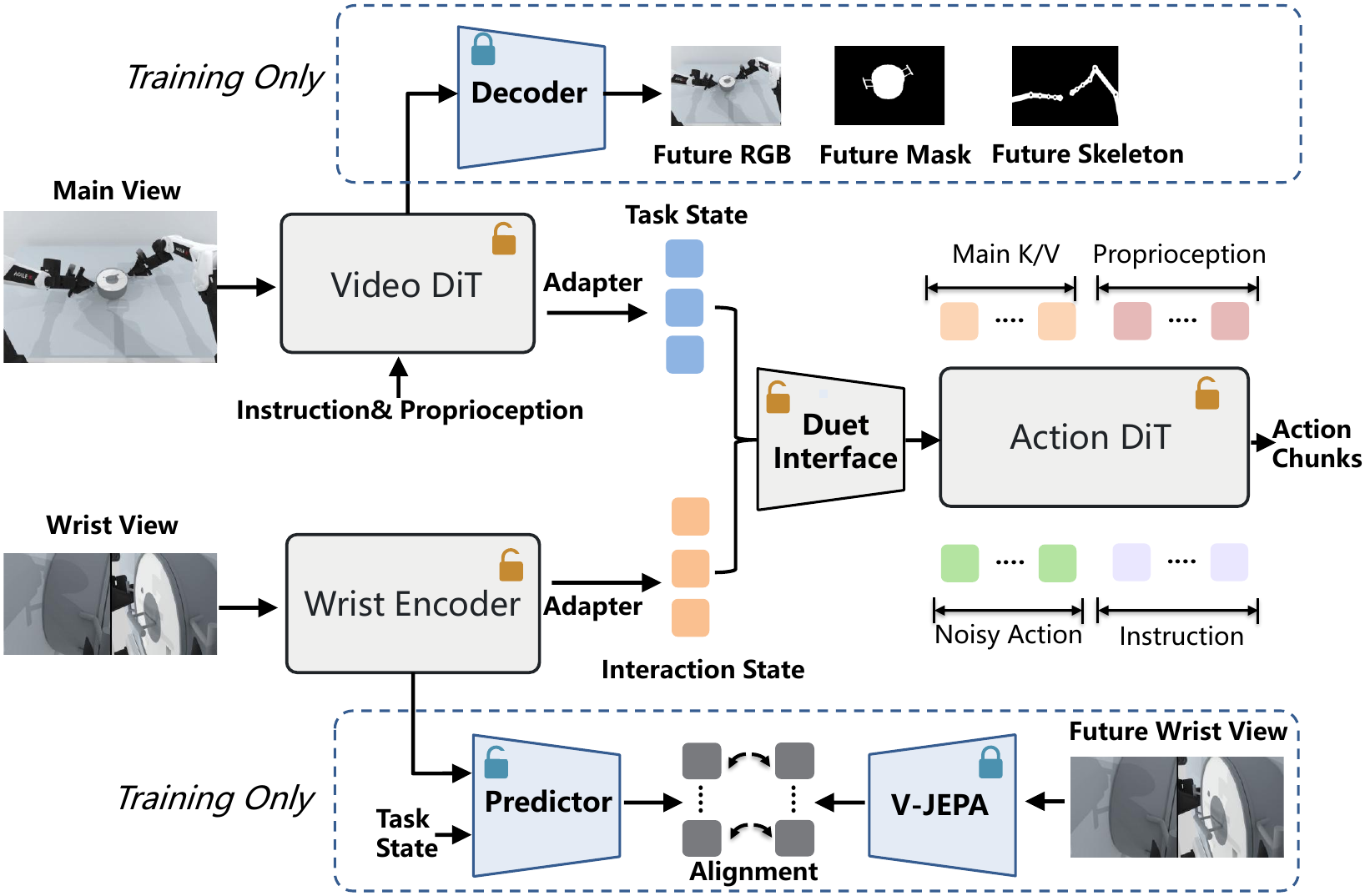}
    \caption{\textbf{Overview of \ours.} Different training-only future objectives shape the Task State and Interaction State. ActionDiT reads both through the Duet Interface; auxiliary predictors are removed at deployment.}
    \label{fig:futureduet-architecture}
\end{figure}

As shown in Fig.~\ref{fig:futureduet-architecture}, VideoDiT extracts main-view features $H^m_t=F_m(o^m_{\leq t},c,p_t)$, and a lightweight online encoder extracts wrist features $H^w_t=E^w_{\omega}(o^w_{\leq t})$. Learned-query adapters summarize these features into a \emph{Task State} and an \emph{Interaction State}:
\begin{equation}
    \begin{aligned}
        S^{\mathrm{task}}_t
        &=A_{\mathrm{task}}(Q^{\mathrm{task}},H^m_t)
        \in\mathbb{R}^{N_{\mathrm{task}}\times d},\\
        S^{\mathrm{int}}_t
        &=A_{\mathrm{int}}(Q^{\mathrm{int}},H^w_t)
        \in\mathbb{R}^{N_{\mathrm{int}}\times d}.
    \end{aligned}
    \label{eq:states}
\end{equation}
Each adapter uses learned queries $Q$ to attend to its stream and produce $N_{\mathrm{task}}$ or $N_{\mathrm{int}}$ tokens of width $d$. The Task State carries scene context from the instruction and proprioception, while the Interaction State summarizes close-range wrist observations.

The Duet Interface projects and fuses the two states for ActionDiT, which also reads current main-view key/value features $K^m_t,V^m_t$~\citep{yuan2026fast,wan2025wan,liang2024mixture}:
\begin{equation}
    \begin{gathered}
        C^{\mathrm{duet}}_t
        =A_{\mathrm{duet}}([S^{\mathrm{task}}_t;S^{\mathrm{int}}_t]),\\
        \widehat v^a_{\tau}
        =\Pi_{\eta}(a^{\tau},\tau;
        C^{\mathrm{duet}}_t,K^m_t,V^m_t,c,p_t).
    \end{gathered}
    \label{eq:action-generation}
\end{equation}
Rather than fusing main and wrist observations at early visual stages, \ours{} defers their combination until each stream has been specialized and summarized into its dedicated state. The Duet Interface projects the Task State and Interaction State into a common conditioning space and supplies the fused tokens to ActionDiT. This interface brings the separately supervised representations together for joint action generation.
Here $[\,;\,]$ denotes token-wise concatenation, and $a^{\tau}$ is a noisy action chunk at flow time $\tau$. ActionDiT generates actions by integrating its predicted velocity from noise. This joint readout combines scene-level task context with local interaction evidence.

\subsection{Task State and Interaction State}
\paragraph{Task State from the main view.}
Future RGB prediction models scene evolution, while interaction masks and robot skeletons focus supervision on task objects and articulated robot motion. The main stream retains the underlying WAM objective: future frames are encoded into VAE latents $z^m$, and VideoDiT learns to predict their flow velocity under Gaussian noise.

A shared DPT-style decoder reads noise-conditioned intermediate VideoDiT features $\widetilde H^{m,\tau}_t$. Separate output heads predict interaction masks and skeleton maps at sampled future offsets $\delta_k$. Through the shared decoder, both auxiliary losses supervise these features, emphasizing object extent and robot configuration alongside the appearance information captured by future RGB prediction. Offline SAM~3~\citep{carion2025sam} masks mark the manipulated object and, when required, its target region; skeleton maps project recorded robot configurations into the calibrated main view. All targets are cached offline. The decoder and output heads are trainable and used only during training; App.~\ref{app:model-details} provides details on interaction-mask generation and robot-skeleton projection.

\paragraph{Interaction State from the wrist views.}
Wrist motion mixes local interaction changes with camera-induced appearance variation. We use future latent prediction to supervise temporal structure without requiring pixel reconstruction. A frozen V-JEPA~2.1~\citep{murlabadia2026vjepa21} encoder processes future wrist clips offline, producing cached targets $Z^w_{t,\mathrm{fut}}$.

Wrist histories provide local evidence but may not reveal the task context governing the next interaction. The Task State supplies this context to a shared training-only predictor:
\begin{equation}
    \widehat Z^w_{t,\mathrm{fut}}
    =P_{\psi}(H^w_t,\operatorname{sg}[S^{\mathrm{task}}_t]).
    \label{eq:wrist-predictor}
\end{equation}
The predictor jointly predicts left- and right-wrist latents, using view embeddings to distinguish the cameras. Future actions are outputs of the policy, not inputs to this predictor. The stop-gradient operator $\operatorname{sg}$ prevents the wrist objective from updating the main stream through the conditioning connection, while allowing it to train the online wrist encoder through $P_{\psi}$. We minimize the mean absolute error
$\mathcal{L}_{\mathrm{wrist}}=|\Omega|^{-1}\sum_{i\in\Omega}|\widehat Z^w_{t,\mathrm{fut},i}-Z^w_{t,\mathrm{fut},i}|$,
where $\Omega$ indexes valid latent elements across the batch and both views, excluding padded future target positions.

\subsection{Joint Action Readout and Training}
\label{sec:joint-training}
Main view RGB prediction and action generation use flow matching. Their noisy inputs are $z^{m,\tau}=(1-\tau)z^m+\tau\epsilon^m$ and $a^{\tau}=(1-\tau)a+\tau\epsilon^a$, with independent standard Gaussian noise. The corresponding objectives are
\begin{equation}
    \begin{aligned}
        \mathcal{L}^{m}_{\mathrm{rgb}}
        &=\mathbb{E}\bigl[
        w(\tau)\|\widehat v^m_{\tau}-(\epsilon^m-z^m)\|_2^2
        \bigr],\\
        \mathcal{L}_{\mathrm{act}}
        &=\mathbb{E}\bigl[
        w(\tau)\|\widehat v^a_{\tau}-(\epsilon^a-a)\|_2^2
        \bigr].
    \end{aligned}
    \label{eq:flow-losses}
\end{equation}
Here $\widehat v^m_{\tau}$ and $\widehat v^a_{\tau}$ are the VideoDiT and ActionDiT velocity predictions. The two objectives use their respective noise-level samples and scheduler weights, denoted by $\tau\in[0,1]$ and $w(\tau)$ for brevity. Mask prediction uses focal--Tversky losses, and skeleton prediction uses focal--Dice losses. We jointly optimize
\begin{equation}
    \begin{split}
        \mathcal{L}={}&\mathcal{L}_{\mathrm{act}}
        +\lambda_{\mathrm{rgb}}\mathcal{L}^{m}_{\mathrm{rgb}}
        +\lambda_{\mathrm{mask}}\mathcal{L}_{\mathrm{mask}}\\
        &+\lambda_{\mathrm{ske}}\mathcal{L}_{\mathrm{ske}}
        +\lambda_{\mathrm{wrist}}\mathcal{L}_{\mathrm{wrist}}.
    \end{split}
    \label{eq:total-loss}
\end{equation}
Main-view future losses train VideoDiT, the wrist objective trains the online wrist encoder, and action loss updates both streams through the Duet Interface. At inference, VideoDiT and the wrist encoder form the two states from observed histories, and ActionDiT reads them jointly as in Eq.~\ref{eq:action-generation}. The VAE decoder, DPT decoder with prediction heads, V-JEPA teacher, and wrist predictor are pruned during evaluation, ensuring that auxiliary future prediction adds no inference overhead. App.~\ref{app:model-details} provides detailed training settings.

\section{Experiments}
\label{sec:experiments}

Our evaluation first measures the complete model across broad bimanual and single-arm benchmarks, then examines three design questions: \textbf{Q1}: Why use a separate wrist pathway? \textbf{Q2}: Why design structured prediction targets for the main view? \textbf{Q3}: And how does this supervision shape the information read by ActionDiT? We answer these questions with benchmark results, controlled component studies, and frozen-checkpoint diagnostics of attention allocation and feature influence.

\subsection{Experimental Setup}
\label{sec:setup}

\paragraph{Benchmarks.}
RoboTwin 2.0~\citep{chen2025robotwin} contains 50 bimanual tasks in clean and randomized settings and serves as our primary benchmark. We jointly train one model on 50 clean and 500 randomized demonstrations per task, totaling 27,500 demonstrations. This setup matches Fast-WAM in data, backbone, preprocessing, training budget, checkpoint selection, and evaluation. Published baseline values come from the cited benchmark reports; specifically, the UWM result follows the RoboTwin50 evaluation in X-WAM~\citep{guo2026xwam}. LIBERO~\citep{liu2023libero} comprises four ten-task, language-conditioned single-arm suites. We use 50 demonstrations per task to evaluate the same design beyond bimanual manipulation.

\paragraph{Controlled studies.}
The main-view factorial study uses six tasks with 500 randomized demonstrations per task. A separate three-task wrist-pathway study uses the same number of demonstrations and reports three-seed means over 100 randomized evaluation episodes per task and seed. The wrist study fixes main-view supervision to RGB to distinguish the effects of pathway separation and future prediction. The factorial study then tests whether structured main-view targets provide complementary gains under both unified and separate wrist pathways. The action-readout analysis uses another six-task set. App.~\ref{app:training-config} provides the task lists, protocols, and training settings. Unless stated otherwise, entries are success rates (SR, \%); ranked tables mark the best and second-best distinct values with green and yellow backgrounds.

\begin{figure}[t]
\centering
\begin{minipage}[t]{0.555\linewidth}
\centering
\captionof{table}{LIBERO benchmark comparison across four single-arm suites. Success rates (\%); higher is better. }
\label{tab:libero}
\vspace{-0.4em}
\scriptsize
\renewcommand{\arraystretch}{0.94}
\setlength{\tabcolsep}{1.1pt}
\resizebox{\linewidth}{!}{%
\begin{tabular}{@{}lccccc@{}}
\toprule
\textbf{Method} & \textbf{Spatial} $\uparrow$ & \textbf{Object} $\uparrow$ & \textbf{Goal} $\uparrow$ & \textbf{Long} $\uparrow$ & \textbf{Avg.} $\uparrow$ \\
\midrule
$\pi_{0}$~\citep{black2024pi0} & 96.8 & 98.8 & 95.8 & 85.2 & 94.1 \\
GR00T-N1~\citep{bjorck2025gr00t} & 94.4 & 97.6 & 93.0 & 90.6 & 93.9 \\
$\pi_{0.5}$~\citep{black2025pi0.5} & 98.6 & 98.2 & 98.0 & 92.4 & 96.8 \\
OpenVLA-OFT~\citep{kim2025openvlaoft} & 97.6 & 98.4 & 97.9 & 94.5 & 97.1 \\
\midrule
Motus~\citep{bi2025motus} & 96.8 & \second{99.8} & 96.6 & 97.6 & 97.7 \\
Fast-WAM~\citep{yuan2026fast} & 98.2 & \best{100.0} & 97.0 & 95.2 & 97.6 \\
MaskWAM~\citep{maskwam} & \second{98.8} & \best{100.0} & \second{98.2} & 96.4 & 98.4 \\
LingBot-VA~\citep{li2026causal} & 98.5 & 99.6 & 97.2 & \best{98.5} & \second{98.5} \\
\midrule
\textbf{\ours} & \best{99.2} & \best{100.0} & \best{99.0} & \second{98.4} & \best{99.2} \\
\bottomrule
\end{tabular}
}
\end{minipage}\hfill
\begin{minipage}[t]{0.425\linewidth}
\centering
\captionof{table}{RoboTwin50 benchmark comparison in clean and randomized settings. }
\label{tab:robotwin-summary}
\vspace{-0.4em}
\scriptsize
\renewcommand{\arraystretch}{0.94}
\setlength{\tabcolsep}{2.0pt}
\begin{tabular*}{\linewidth}{@{\extracolsep{\fill}}lrr@{}}
\toprule
\textbf{Method} & \textbf{Clean} $\uparrow$ & \textbf{Rand.} $\uparrow$ \\
\midrule
$\pi_{0}$~\citep{black2024pi0} & 65.9 & 58.4 \\
$X$-VLA~\citep{zheng2025xvla} & 72.9 & 72.8 \\
$\pi_{0.5}$~\citep{black2025pi0.5} & 82.7 & 76.8 \\
\addlinespace[1pt]
UWM~\citep{zhu2025unifiedworldmodelscoupling} & 81.7 & 78.6 \\
Motus~\citep{bi2025motus} & 88.7 & 87.0 \\
$X$-WAM~\citep{guo2026xwam} & 89.8 & 90.7 \\
Fast-WAM~\citep{yuan2026fast} & 91.9 & \second{91.8} \\
LingBot-VA~\citep{li2026causal} & \second{92.9} & 91.5 \\
\midrule
\textbf{\ours} & \best{94.2} & \best{94.1} \\
\bottomrule
\end{tabular*}
\end{minipage}
\end{figure}

\begin{figure}[!h]
  \centering
  \includegraphics[width=\linewidth]{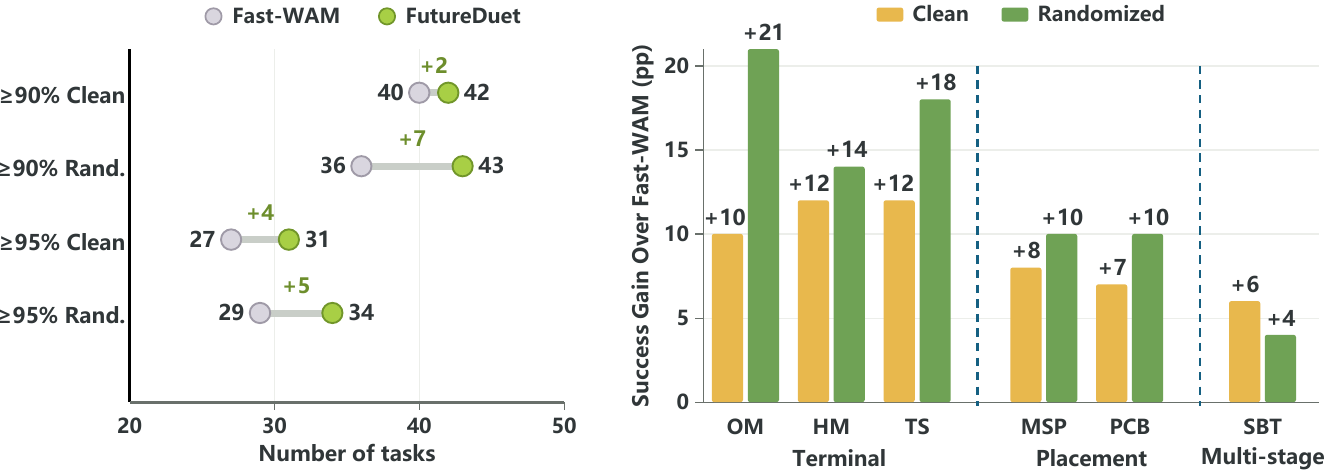}
  \caption{\textbf{RoboTwin50 performance varies substantially across tasks.} (a) Number of tasks reaching at least 90\% and 95\% success under clean and randomized evaluation. (b) Gains over Fast-WAM on the six tasks that remain below 90\% for \ours\ in both settings. Gains are measured in percentage points (pp). OM, HM, TS, MSP, PCB, and SBT denote Open Microwave, Hanging Mug, Turn Switch, Move Stapler Pad, Place Can Basket, and Stack Bowls Three, respectively.}
  \label{fig:robotwin-landscape}
\end{figure}

\subsection{Overall Performance}
\label{sec:main-results}
\raggedbottom

\paragraph{LIBERO.}
Tab.~\ref{tab:libero} compares \ours{} with source-reported results across four single-arm suites. \ours{} achieves 99.2\% average success, with the best reported results on Spatial, Object, and Goal, demonstrating its effectiveness beyond the bimanual RoboTwin setting.

\paragraph{RoboTwin50.}
\ours{} achieves 94.2\% success in the clean setting and 94.1\% under randomization, surpassing the reported Fast-WAM results by 2.3 percentage points in both settings (Tab.~\ref{tab:robotwin-summary}).

\subsection{Task-Level Analysis on RoboTwin50}

\paragraph{Performance across tasks.}
The high benchmark average coexists with a small set of challenging tasks (Fig.~\ref{fig:robotwin-landscape}). \ours\ reaches at least 90\% success on 41 of 50 tasks in both settings, while six remain below this threshold in both. These tasks require precise terminal interaction, constrained placement, or precision across multiple stages. On this subset, \ours\ improves over Fast-WAM by 9.2 percentage points in clean evaluation and 12.8 percentage points under randomization, compared with 2.3 points benchmark-wide. Complete results and task groupings appear in App.~\ref{app:robotwin-results}.

\begin{figure}[!h]
  \centering
  \includegraphics[width=\linewidth]{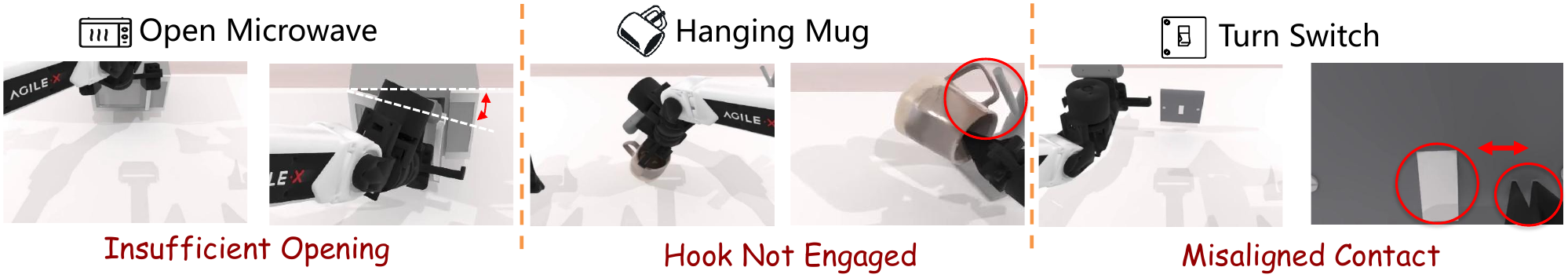}
  \caption{\textbf{Failure cases on three challenging interaction tasks.} The policy reaches the relevant interaction region but fails to complete the required opening, engagement, or contact.}
  \label{fig:fine-grained-failures}
\end{figure}

\Needspace{9\baselineskip}
\paragraph{Gains on challenging interactions.}
The six remaining challenging tasks account for 48\% of the total net success gain over Fast-WAM in clean evaluation and 66\% under randomization. Open Microwave, Hanging Mug, and Turn Switch show the largest gains within this group in both settings. Their illustrated failures share a common structure: the policy reaches the relevant object but does not complete the final local transition---the door must open far enough, the mug handle must engage the hook, and the end effector must align with and actuate the switch (Fig.~\ref{fig:fine-grained-failures}). These examples suggest that success depends not only on reaching the interaction region, but also on resolving and executing the contact geometry required for completion. Because wrist views provide close-range evidence of these terminal transitions, these cases motivate the controlled wrist-pathway study in Sec.~\ref{sec:design-analysis}, where we isolate the contributions of wrist access, pathway separation, and future supervision on the same three tasks.

\vspace{-0.5em}
\subsection{Why Use a Separate Wrist Pathway?}
\label{sec:design-analysis}

\noindent
\begin{minipage}[t]{0.48\linewidth}
\vspace{0pt}
We isolate three factors on the wrist side: observation access, pathway separation, and future supervision (Fig.~\ref{fig:wrist-efficiency}). Main-view supervision remains RGB-only throughout this study. We first compare unified and separate wrist pathways, then hold the lightweight pathway fixed to evaluate different prediction targets.

\paragraph{Wrist access and pathway separation.}
Incorporating wrist observations into the unified pathway boosts mean success from 34.3\% to 55.8\%, validating the value of close-range wrist cues. Remarkably, a lightweight separate pathway reaches 60.9\% even without auxiliary wrist future prediction, demonstrating the intrinsic benefit of a decoupled stream.
\end{minipage}\hfill
\begin{minipage}[t]{0.5\linewidth}
\vspace{5pt}
\centering
\includegraphics[width=\linewidth]{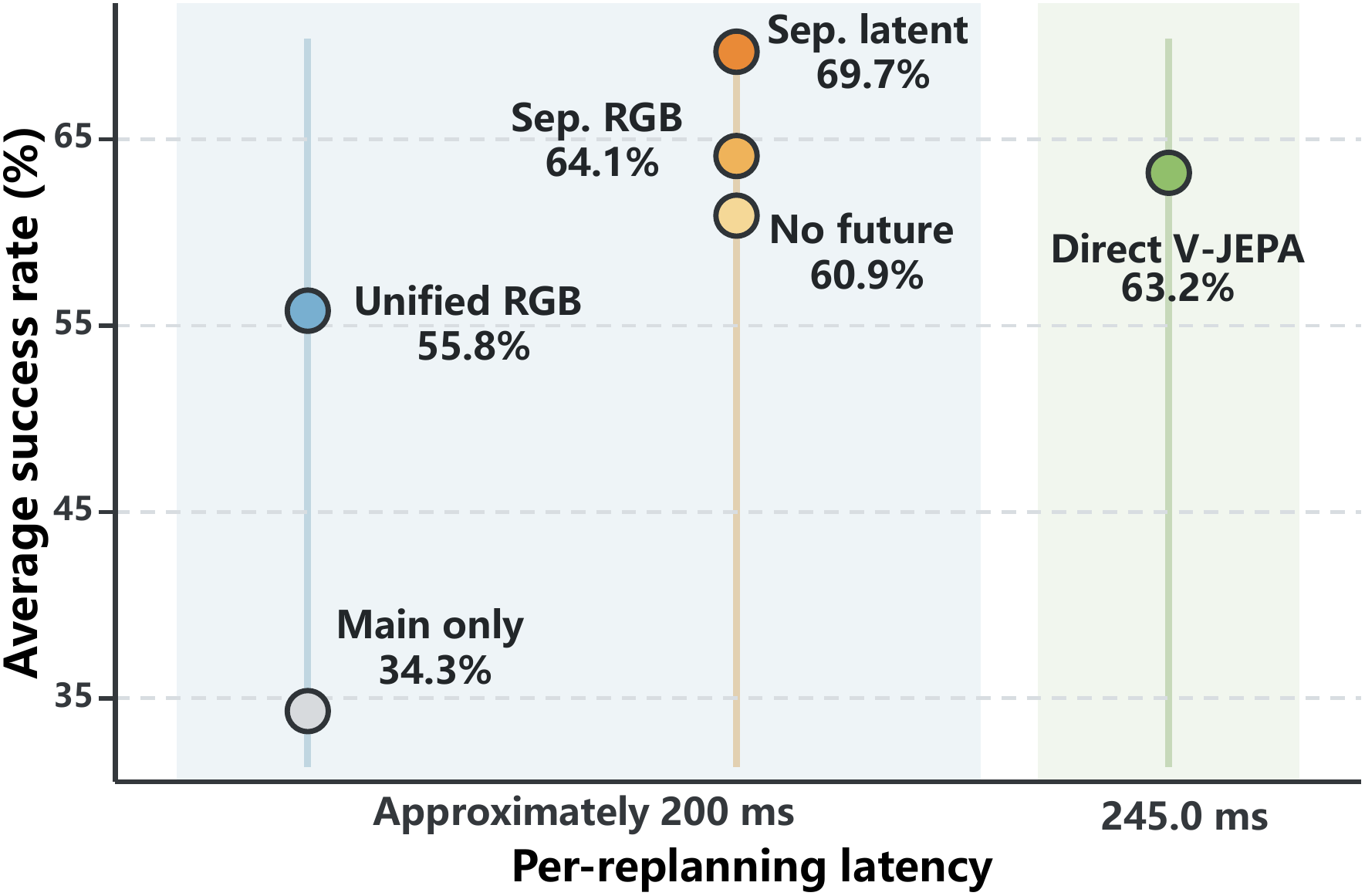}
\captionof{figure}{Three-seed wrist-pathway study with RGB-only main-view supervision.}
\label{fig:wrist-efficiency}
\end{minipage}

\paragraph{Future supervision under a fixed pathway.}
Applying future RGB supervision boosts success from 60.9\% to 64.1\%, whereas future-latent supervision further elevates performance to 69.7\%. The RGB result confirms that wrist future prediction is also beneficial, rather than being restricted to latent targets. Nevertheless, the additional 5.6-point gain from latent supervision demonstrates that the choice of target representation is critical within the same pathway. While both objectives provide auxiliary temporal signals beyond action loss, the latent target effectively leverages rich pretrained video features without demanding pixel-level reconstruction. Overall, these findings validate pairing a decoupled wrist pathway with future-latent prediction, achieving optimal performance while preserving deployment efficiency.

\paragraph{Deployment efficiency.}
Deploying a heavy encoder like V-JEPA highlights the trade-off between model capacity and efficiency, increasing latency from 196.86~ms to 244.99~ms for a 63.2\% success rate. We bypass this limitation through training-only auxiliary supervision: confining future-latent prediction strictly to training enables our lightweight pathway to achieve 69.7\% success at near-baseline latency ($\sim$200~ms). Since all auxiliary targets share the same deployed execution graph, upgrading the training target enriches feature representations with zero inference overhead.

\vspace{-0.5em}
\subsection{Designing Structured Supervision for the Main View }
\label{sec:level-contributions}

Decoupling future supervision allows us to design the main and wrist pathways around the information they provide for control. 
With the wrist pathway established, we now investigate supervision strategies for the main view. Future RGB prediction captures scene evolution but does not explicitly prioritize the task objects and robot motion that guide manipulation. We therefore study whether masks and skeletons strengthen this supervision and complement the separate wrist pathway (Tab.~\ref{tab:ablation-six}).

\begin{table}[t]
\centering
\caption{\textbf{Factorial study of main-view supervision and wrist pathways.}
The study crosses four main-view target configurations with unified future-RGB and separate future-latent wrist pathways. Fine-grained execution comprises Open Microwave (OM), Hanging Mug (HM), and Turn Switch (TS); additional tasks comprise Adjust Bottle (AB), Blocks Ranking RGB (BR), and Grab Roller (GR). Green and yellow shading indicate the best and second-best result in each column.}
\label{tab:ablation-six}
\small
\renewcommand{\arraystretch}{1.02}
\setlength{\tabcolsep}{4.0pt}
\begin{tabularx}{\linewidth}{@{}l>{\raggedright\arraybackslash}X*{7}{c}@{}}
\toprule
\multicolumn{2}{c}{\textbf{Predictive configuration}} &
\multicolumn{3}{c}{\textbf{Fine-grained execution}} &
\multicolumn{3}{c}{\textbf{Additional tasks}} &
\multicolumn{1}{c}{\textbf{Overall}} \\
\cmidrule(lr){1-2}
\cmidrule(lr){3-5}
\cmidrule(lr){6-8}
\cmidrule(lr){9-9}
\textbf{Main-view targets} &
\textbf{Wrist pathway} &
\textbf{OM} &
\textbf{HM} &
\textbf{TS} &
\textbf{AB} &
\textbf{BR} &
\textbf{GR} &
\textbf{Avg.} \\
\midrule
RGB &
Unified (future RGB) &
43 & 60 & 57 & 88 & 90 & 88 & 71.0 \\

RGB+Mask &
Unified (future RGB) &
43 & 60 & 58 & \best{92} & \second{93} & 89 & 72.5 \\

RGB+Skel. &
Unified (future RGB) &
44 & 61 & 59 & 89 & 91 & \best{92} & 72.7 \\

RGB+Mask+Skel. &
Unified (future RGB) &
45 & 62 & 59 & \second{91} & \best{94} & \second{90} & 73.5 \\
\midrule
RGB &
Separate (future latent) &
59 & 66 & 68 & 88 & 88 & 87 & 76.0 \\

RGB+Mask &
Separate (future latent) &
\second{60} & 66 & \second{69} &
\best{92} & \second{93} & 87 & \second{77.8} \\

RGB+Skel. &
Separate (future latent) &
\second{60} & \second{68} & \best{70} &
88 & 89 & \second{90} & 77.5 \\

\textbf{RGB+Mask+Skel.} &
\textbf{Separate (future latent)} &
\best{65} & \best{70} & \best{70} &
\second{91} & \best{94} & \best{92} & \best{80.3} \\
\bottomrule
\end{tabularx}
\par\vspace{1mm}

\end{table}
\paragraph{Complementarity with wrist modeling.}
Adding masks and skeletons raises average success from 71.0\% to 73.5\% under the unified pathway and from 76.0\% to 80.3\% under the separate future-latent pathway, with gains spanning both fine-grained execution and the additional tasks. 
With RGB-only main-view supervision, the gains from the separate future-latent pathway are concentrated on OM, HM, and TS. Adding structured main-view targets also improves AB, BR, and GR under both pathways, broadening the benefits beyond the fine-grained tasks most responsive to wrist modeling. These improvements show that the value of separately designing future supervision extends beyond the wrist pathway. Alongside wrist-side prediction, explicitly emphasizing task objects and robot motion in the main-view objective provides additional gains, with the combined design achieving the highest average success under both wrist pathways.

\paragraph{Distinct contributions of masks and skeletons.}
Under both wrist pathways, masks provide larger gains than skeletons on Adjust Bottle and Blocks Ranking RGB, whereas skeletons provide larger gains on Grab Roller. Neither target is consistently better across these tasks, and combining them achieves the highest average success under either pathway. This pattern supports supervising both task objects and robot motion rather than relying on either target alone.

Together, these results show that structured main view supervision complements wrist modeling, with masks and skeletons contributing differently across tasks. We next examine whether this functional division is mirrored in how ActionDiT allocates attention to task objects and robot features.

\subsection{Understanding the Effects of Structured Prediction Targets}
\label{sec:diagnostics}

The factorial study shows that masks and skeletons produce different task-level gains. We therefore examine whether they change the main-view evidence used by ActionDiT. Across 48 matched queries from separately trained six-task checkpoints, we first use action-to-visual attention to measure where action queries retrieve information, and then apply matched K/V interventions to test whether features from those regions affect the action output.

\paragraph{Where does ActionDiT read?}
We extract action-query-to-main-view attention from the final two joint VideoDiT--ActionDiT attention blocks. RR(O) and RR(G) quantify action attention directed to task-object and executing-robot regions relative to a size-matched background, whereas OCE measures object-center alignment, with lower values indicating better alignment.

\begin{figure}[h]
  \centering
  \includegraphics[width=0.84\linewidth]{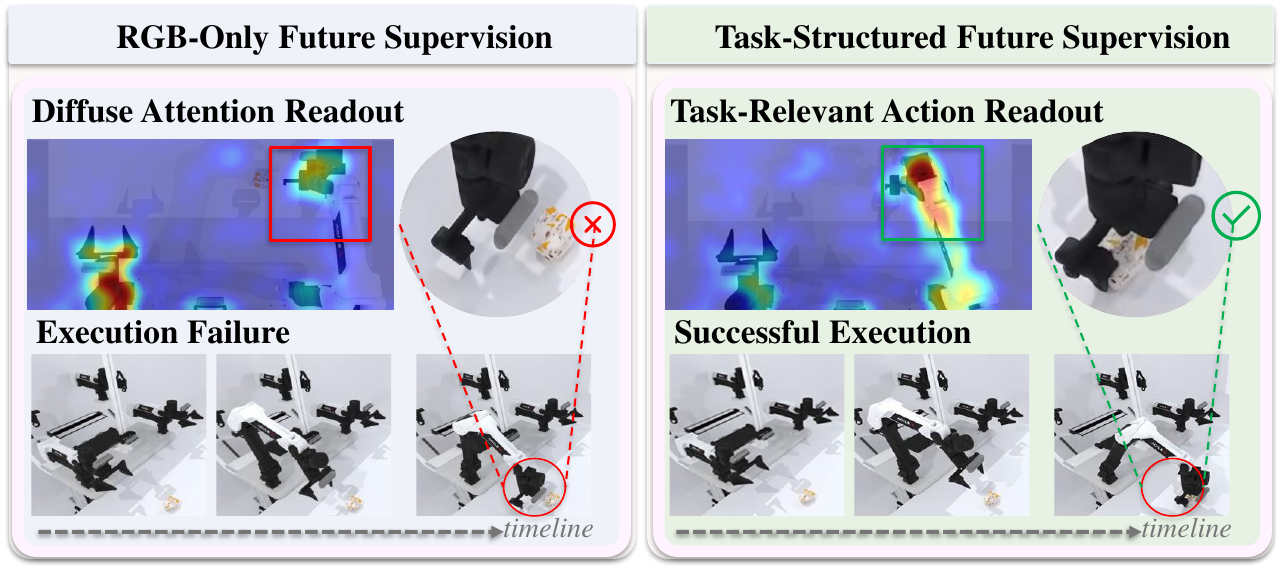}
  \caption{\textbf{Structured targets focus action readout.} Compared with RGB-only future supervision, joint mask-and-skeleton supervision focuses readout on the task-relevant interaction.}
  \label{fig:structured-readout-qualitative}
\end{figure}

The effects are target-specific. Under the separate wrist pathway, mask supervision raises object readout RR(O) from 1.10 to 1.46 and improves object-center alignment, whereas skeleton supervision produces the strongest robot-side change, raising RR(G) from 1.20 to 2.00. The same division appears under unified wrist modeling, indicating that it is induced by the main-view targets rather than the wrist pathway. Fig.~\ref{fig:structured-readout-qualitative} qualitatively illustrates this effect, showing how joint supervision concentrates action readout on the task-relevant interaction.

\noindent
\begin{minipage}[t]{0.57\linewidth}
\vspace{0pt}
\paragraph{Which regions influence action generation?}
Attention indicates where action queries retrieve information, but not whether the retrieved features affect the output. We therefore replace object- or robot-region K/V features with the mean main-view K/V and measure the resulting action change using R(O) and R(G).

The intervention results mirror the attention patterns. Mask supervision primarily increases sensitivity to object features, whereas skeleton supervision primarily increases sensitivity to robot features; joint supervision preserves both effects under either wrist pathway. Thus, the target-specific readout is reflected not only in attention patterns but also in the action model's response to the corresponding visual features.
\end{minipage}\hfill
\begin{minipage}[t]{0.39\linewidth}
\vspace{0pt}
\centering
\includegraphics[width=\linewidth]{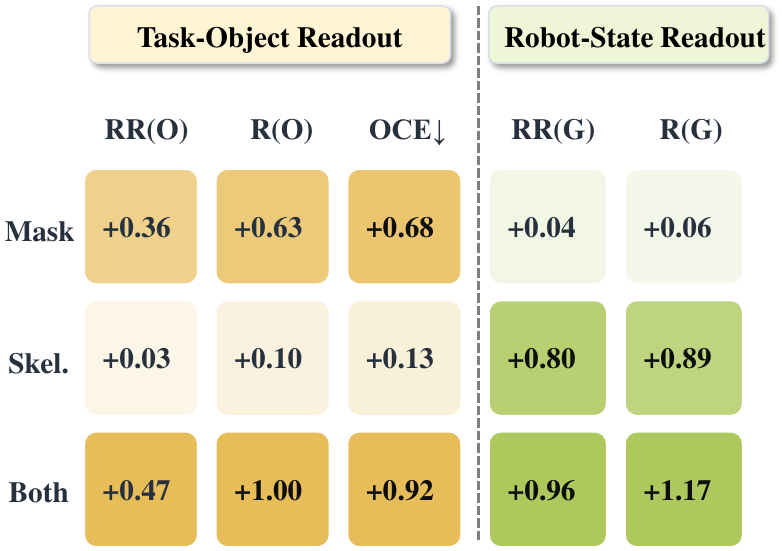}
\captionof{figure}{\textbf{Target-specific changes under separate wrist modeling.} Positive values indicate increases in attention readout and feature sensitivity, as well as reductions in OCE.}
\label{fig:structured-readout-quantitative}
\end{minipage}

\medskip
Together, the two diagnostics provide complementary evidence: attention identifies where ActionDiT reads, while K/V intervention shows that the corresponding features affect its output. Their agreement suggests that mask and skeleton supervision shape complementary sides of the Task State---the task object and the robot executing the motion---helping explain why their combination performs best in the factorial study. Full results appear in Appendix Tab.~\ref{tab:app-action-readout-full}.
\section{Conclusion}
\label{sec:conclusion}

\ours{} decouples two choices that existing WAMs often entangle together: which observations support control and what future each visual stream is trained to predict. By using main and wrist observations for action generation while predicting structured main-view targets and future wrist latents, our framework learns complementary representations of overall task progress and local interaction. This decoupled design improves performance across RoboTwin50 and LIBERO, with the largest RoboTwin50 gains appearing on tasks requiring precise interaction completion. Controlled studies isolate two key sources of wrist-side improvement: pathway separation provides a fundamental architectural benefit, while supervision target selection favors future-latent prediction over pixel-level RGB reconstruction under the same deployed architecture. Meanwhile, structured main-view supervision offers complementary gains.
While our evaluation is currently limited to simulation, future work will test \ours{} on real robots and explore broader view-target configurations. Overall, our results demonstrate that distinct camera streams can jointly guide action control without forcing every view to predict a uniform future, highlighting the value of decoupling observation access from training-time auxiliary future supervision in world action models.

\bibliography{refs}

\begin{thebibliography}{51}
\providecommand{\natexlab}[1]{#1}
\providecommand{\url}[1]{\texttt{#1}}
\expandafter\ifx\csname urlstyle\endcsname\relax
  \providecommand{\doi}[1]{doi: #1}\else
  \providecommand{\doi}{doi: \begingroup \urlstyle{rm}\Url}\fi

\bibitem[An et~al.(2026)An, Hu, Huang, Huang, Li, Liang, Shao, Song, Wang,
  Yuan, Zhang, Zhang, Zhuang, and Li]{aiflowPerspective}
Hongjun An, Wenhan Hu, Sida Huang, Siqi Huang, Ruanjun Li, Yuanzhi Liang,
  Jiawei Shao, Yiliang Song, Zihan Wang, Cheng Yuan, Chi Zhang, Hongyuan Zhang,
  Wenhao Zhuang, and Xuelong Li.
\newblock {AI Flow}: Perspectives, scenarios, and approaches.
\newblock \emph{Vicinagearth}, 3\penalty0 (1):\penalty0 1, 2026.
\newblock \doi{10.1007/s44336-025-00031-y}.

\bibitem[Bharadhwaj et~al.(2024)Bharadhwaj, Mottaghi, Gupta, and
  Tulsiani]{bharadhwaj2024track2act}
Homanga Bharadhwaj, Roozbeh Mottaghi, Abhinav Gupta, and Shubham Tulsiani.
\newblock Track2act: Predicting point tracks from internet videos enables
  generalizable robot manipulation.
\newblock In \emph{European Conference on Computer Vision}, pp.\  306--324.
  Springer, 2024.

\bibitem[Bharadhwaj et~al.(2025)Bharadhwaj, Dwibedi, Gupta, Tulsiani, Doersch,
  Xiao, Shah, Xia, Sadigh, and Kirmani]{bharadhwaj2024gen2act}
Homanga Bharadhwaj, Debidatta Dwibedi, Abhinav Gupta, Shubham Tulsiani, Carl
  Doersch, Ted Xiao, Dhruv Shah, Fei Xia, Dorsa Sadigh, and Sean Kirmani.
\newblock {Gen2Act}: Human video generation in novel scenarios enables
  generalizable robot manipulation.
\newblock In \emph{Proceedings of the 9th Conference on Robot Learning}, volume
  305 of \emph{Proceedings of Machine Learning Research}, pp.\  3936--3951.
  PMLR, 2025.
\newblock URL \url{https://proceedings.mlr.press/v305/bharadhwaj25a.html}.

\bibitem[Bi et~al.(2025)Bi, Tan, Xie, Wang, Huang, Liu, Zhao, Feng, Xiang,
  Rong, Zhao, Liu, Su, Ma, Su, and Zhu]{bi2025motus}
Hongzhe Bi, Hengkai Tan, Shenghao Xie, Zeyuan Wang, Shuhe Huang, Haitian Liu,
  Ruowen Zhao, Yao Feng, Chendong Xiang, Yinze Rong, Hongyan Zhao, Hanyu Liu,
  Zhizhong Su, Lei Ma, Hang Su, and Jun Zhu.
\newblock Motus: A unified latent action world model.
\newblock \emph{arXiv preprint arXiv:2512.13030}, 2025.

\bibitem[Black et~al.(2025)Black, Brown, Driess, Esmail, Equi, Finn, Fusai,
  Groom, Hausman, Ichter, Jakubczak, Jones, Ke, Levine, Li-Bell, Mothukuri,
  Nair, Pertsch, Shi, Smith, Tanner, Vuong, Walling, Wang, and
  Zhilinsky]{black2024pi0}
Kevin Black, Noah Brown, Danny Driess, Adnan Esmail, Michael~Robert Equi,
  Chelsea Finn, Niccolo Fusai, Lachy Groom, Karol Hausman, Brian Ichter, Szymon
  Jakubczak, Tim Jones, Liyiming Ke, Sergey Levine, Adrian Li-Bell, Mohith
  Mothukuri, Suraj Nair, Karl Pertsch, Lucy~Xiaoyang Shi, Laura Smith, James
  Tanner, Quan Vuong, Anna Walling, Haohuan Wang, and Ury Zhilinsky.
\newblock $\pi_0$: A vision-language-action flow model for general robot
  control.
\newblock In \emph{Proceedings of Robotics: Science and Systems}, Los Angeles,
  CA, USA, June 2025.
\newblock \doi{10.15607/RSS.2025.XXI.010}.

\bibitem[Brohan et~al.(2023)]{brohan2023rt2}
Anthony Brohan et~al.
\newblock {RT-2}: Vision-language-action models transfer web knowledge to
  robotic control.
\newblock \emph{arXiv preprint arXiv:2307.15818}, 2023.
\newblock URL \url{https://arxiv.org/abs/2307.15818}.

\bibitem[Carion et~al.(2025)Carion, Gustafson, Hu, Debnath, Hu, Suris, Ryali,
  Alwala, Khedr, Huang, et~al.]{carion2025sam}
Nicolas Carion, Laura Gustafson, Yuan-Ting Hu, Shoubhik Debnath, Ronghang Hu,
  Didac Suris, Chaitanya Ryali, Kalyan~Vasudev Alwala, Haitham Khedr, Andrew
  Huang, et~al.
\newblock {SAM 3}: Segment anything with concepts.
\newblock \emph{arXiv preprint arXiv:2511.16719}, 2025.

\bibitem[Cen et~al.(2025{\natexlab{a}})Cen, Huang, Yuan, Yuan, Yu, Jiang, Guo,
  Li, Luo, Wang, Li, Zhao, and Chen]{cen2025rynnvla}
Jun Cen, Siteng Huang, Yuqian Yuan, Hangjie Yuan, Chaohui Yu, Yuming Jiang,
  Jiayan Guo, Kehan Li, Hao Luo, Fan Wang, Xin Li, Deli Zhao, and Hao Chen.
\newblock Rynnvla-002: A unified vision-language-action and world model.
\newblock \emph{arXiv preprint arXiv:2511.17502}, 2025{\natexlab{a}}.

\bibitem[Cen et~al.(2025{\natexlab{b}})Cen, Yu, Yuan, Jiang, Huang, Guo, Li,
  Song, Luo, Wang, Zhao, and Chen]{cen2025worldvla}
Jun Cen, Chaohui Yu, Hangjie Yuan, Yuming Jiang, Siteng Huang, Jiayan Guo, Xin
  Li, Yibing Song, Hao Luo, Fan Wang, Deli Zhao, and Hao Chen.
\newblock {WorldVLA}: Towards autoregressive action world model.
\newblock \emph{arXiv preprint arXiv:2506.21539}, 2025{\natexlab{b}}.

\bibitem[Cheang et~al.(2024)Cheang, Chen, Jing, Kong, Li, Li, Liu, Wu, Xu,
  Yang, Zhang, and Zhu]{cheang2024gr2generativevideolanguageactionmodel}
Chi-Lam Cheang, Guangzeng Chen, Ya~Jing, Tao Kong, Hang Li, Yifeng Li, Yuxiao
  Liu, Hongtao Wu, Jiafeng Xu, Yichu Yang, Hanbo Zhang, and Minzhao Zhu.
\newblock Gr-2: A generative video-language-action model with web-scale
  knowledge for robot manipulation.
\newblock \emph{arXiv preprint arXiv:2410.06158}, 2024.

\bibitem[Chen et~al.(2026{\natexlab{a}})Chen, Wang, Chen, Chen, Gao, Tang, Li,
  Liu, Yao, Li, Xu, and Yu]{chen2026lawam}
Jialei Chen, Kai Wang, Kang Chen, Shuaihang Chen, Feng Gao, Wenhao Tang,
  Zhiyuan Li, Weilin Liu, Zhuyu Yao, Boxun Li, Yuanbo Xu, and Chao Yu.
\newblock {LaWAM}: Latent world action models for efficient dynamics-aware
  robot policies.
\newblock \emph{arXiv preprint arXiv:2606.15768}, 2026{\natexlab{a}}.
\newblock URL \url{https://arxiv.org/abs/2606.15768}.

\bibitem[Chen et~al.(2026{\natexlab{b}})Chen, Chen, Chen, Cai, Liu, Li, Liang,
  Lin, Ge, Gu, Deng, Guo, Nian, Xie, Chen, Su, Xu, Liu, Hu, Gao, Wang, Liang,
  Qin, Yang, Luo, and Mu]{chen2025robotwin}
Tianxing Chen, Zanxin Chen, Baijun Chen, Zijian Cai, Yibin Liu, Zixuan Li,
  Qiwei Liang, Xianliang Lin, Yiheng Ge, Zhenyu Gu, Weiliang Deng, Yubin Guo,
  Tian Nian, Xuanbing Xie, Qiangyu Chen, Kailun Su, Tianling Xu, Guodong Liu,
  Mengkang Hu, Huan-ang Gao, Kaixuan Wang, Zhixuan Liang, Yusen Qin, Xiaokang
  Yang, Ping Luo, and Yao Mu.
\newblock {RoboTwin 2.0}: A scalable data generator and benchmark with strong
  domain randomization for robust bimanual robotic manipulation.
\newblock In \emph{Proceedings of the 43rd International Conference on Machine
  Learning}, 2026{\natexlab{b}}.
\newblock URL \url{https://icml.cc/virtual/2026/poster/62192}.

\bibitem[Du et~al.(2023)Du, Yang, Dai, Dai, Nachum, Tenenbaum, Schuurmans, and
  Abbeel]{du2023learninguniversalpoliciestextguided}
Yilun Du, Mengjiao Yang, Bo~Dai, Hanjun Dai, Ofir Nachum, Joshua~B. Tenenbaum,
  Dale Schuurmans, and Pieter Abbeel.
\newblock Learning universal policies via text-guided video generation.
\newblock In \emph{Advances in Neural Information Processing Systems},
  volume~36, 2023.
\newblock \doi{10.52202/075280-0403}.
\newblock URL
  \url{https://proceedings.neurips.cc/paper_files/paper/2023/hash/1d5b9233ad716a43be5c0d3023cb82d0-Abstract-Conference.html}.

\bibitem[Feng et~al.(2025)Feng, Tan, Mao, Liu, Huang, Xiang, Su, and
  Zhu]{feng2025vidarembodiedvideodiffusion}
Yao Feng, Hengkai Tan, Xinyi Mao, Guodong Liu, Shuhe Huang, Chendong Xiang,
  Hang Su, and Jun Zhu.
\newblock Vidar: Embodied video diffusion model for generalist bimanual
  manipulation, 2025.
\newblock URL \url{https://arxiv.org/abs/2507.12898}.

\bibitem[Guo et~al.(2026)Guo, Li, Li, Chen, Sun, Su, Wang, Zhang, Li, and
  Liu]{guo2026xwam}
Jun Guo, Qiwei Li, Peiyan Li, Zilong Chen, Nan Sun, Yifei Su, Heyun Wang, Yuan
  Zhang, Xinghang Li, and Huaping Liu.
\newblock Unified {4D} world action modeling from video priors with
  asynchronous denoising, 2026.
\newblock URL \url{https://arxiv.org/abs/2604.26694}.
\newblock arXiv preprint arXiv:2604.26694.

\bibitem[Haldar \& Pinto(2025)Haldar and Pinto]{haldar2025point}
Siddhant Haldar and Lerrel Pinto.
\newblock Point policy: Unifying observations and actions with key points for
  robot manipulation.
\newblock \emph{arXiv preprint arXiv:2502.20391}, 2025.

\bibitem[Hu et~al.(2025)Hu, Guo, Wang, Chen, Wang, Zhang, Sreenath, Lu, and
  Chen]{hu2024video}
Yucheng Hu, Yanjiang Guo, Pengchao Wang, Xiaoyu Chen, Yen-Jen Wang, Jianke
  Zhang, Koushil Sreenath, Chaochao Lu, and Jianyu Chen.
\newblock Video prediction policy: A generalist robot policy with predictive
  visual representations.
\newblock In \emph{Proceedings of the 42nd International Conference on Machine
  Learning}, volume 267 of \emph{Proceedings of Machine Learning Research},
  pp.\  24328--24346. PMLR, 2025.
\newblock URL \url{https://proceedings.mlr.press/v267/hu25g.html}.

\bibitem[Huang et~al.(2026)Huang, Wu, Zhang, Wang, You, and
  Huang]{huang2026forewam}
Jiakai Huang, Zhongbo Wu, Zheng Zhang, Zihan Wang, Shan You, and Tao Huang.
\newblock Foresight without seeing: Latent futures for world action models.
\newblock \emph{arXiv preprint arXiv:2608.11605}, 2026.
\newblock URL \url{https://arxiv.org/abs/2608.11605}.

\bibitem[Jang et~al.(2025)Jang, Ye, Lin, Xiang, Bjorck, Fang, Hu, Huang,
  Kundalia, Lin, Magne, Mandlekar, Narayan, Tan, Wang, Wang, Wang, Xu, Zeng,
  Zheng, Zheng, Liu, Zettlemoyer, Fox, Kautz, Reed, Zhu, and
  Fan]{jang2025dreamgenunlockinggeneralizationrobot}
Joel Jang, Seonghyeon Ye, Zongyu Lin, Jiannan Xiang, Johan Bjorck, Yu~Fang,
  Fengyuan Hu, Spencer Huang, Kaushil Kundalia, Yen-Chen Lin, Loic Magne, Ajay
  Mandlekar, Avnish Narayan, You~Liang Tan, Guanzhi Wang, Jing Wang, Qi~Wang,
  Yinzhen Xu, Xiaohui Zeng, Kaiyuan Zheng, Ruijie Zheng, Ming-Yu Liu, Luke
  Zettlemoyer, Dieter Fox, Jan Kautz, Scott Reed, Yuke Zhu, and Linxi Fan.
\newblock Dreamgen: Unlocking generalization in robot learning through video
  world models, 2025.
\newblock URL \url{https://arxiv.org/abs/2505.12705}.

\bibitem[Kim et~al.(2025{\natexlab{a}})Kim, Finn, and Liang]{kim2025openvlaoft}
Moo~Jin Kim, Chelsea Finn, and Percy Liang.
\newblock Fine-tuning vision-language-action models: Optimizing speed and
  success.
\newblock In \emph{Proceedings of Robotics: Science and Systems}, Los Angeles,
  CA, USA, June 2025{\natexlab{a}}.
\newblock \doi{10.15607/RSS.2025.XXI.017}.

\bibitem[Kim et~al.(2025{\natexlab{b}})Kim, Pertsch, Karamcheti, Xiao,
  Balakrishna, Nair, Rafailov, Foster, Sanketi, Vuong, Kollar, Burchfiel,
  Tedrake, Sadigh, Levine, Liang, and Finn]{kim2024openvla}
Moo~Jin Kim, Karl Pertsch, Siddharth Karamcheti, Ted Xiao, Ashwin Balakrishna,
  Suraj Nair, Rafael Rafailov, Ethan~P. Foster, Pannag~R. Sanketi, Quan Vuong,
  Thomas Kollar, Benjamin Burchfiel, Russ Tedrake, Dorsa Sadigh, Sergey Levine,
  Percy Liang, and Chelsea Finn.
\newblock {OpenVLA}: An open-source vision-language-action model.
\newblock In \emph{Proceedings of the 8th Conference on Robot Learning}, volume
  270 of \emph{Proceedings of Machine Learning Research}, pp.\  2679--2713.
  PMLR, 2025{\natexlab{b}}.
\newblock URL \url{https://proceedings.mlr.press/v270/kim25c.html}.

\bibitem[Kim et~al.(2026)Kim, Gao, Lin, Lin, Ge, Lam, Liang, Song, Liu, Finn,
  and Gu]{kim2026cosmos}
Moo~Jin Kim, Yihuai Gao, Tsung-Yi Lin, Yen-Chen Lin, Yunhao Ge, Grace Lam,
  Percy Liang, Shuran Song, Ming-Yu Liu, Chelsea Finn, and Jinwei Gu.
\newblock Cosmos policy: Fine-tuning video models for visuomotor control and
  planning.
\newblock \emph{arXiv preprint arXiv:2601.16163}, 2026.

\bibitem[Li et~al.(2026)Li, Zhang, Luo, Yang, Wang, Han, Yu, Gao, Xue, Zhu,
  et~al.]{li2026causal}
Lin Li, Qihang Zhang, Yiming Luo, Shuai Yang, Ruilin Wang, Fei Han, Mingrui Yu,
  Zelin Gao, Nan Xue, Xing Zhu, et~al.
\newblock Causal world modeling for robot control.
\newblock \emph{arXiv preprint arXiv:2601.21998}, 2026.

\bibitem[Liang et~al.(2025)Liang, Tokmakov, Liu, Sudhakar, Shah, Ambrus, and
  Vondrick]{liang2025videogenerators}
Junbang Liang, Pavel Tokmakov, Ruoshi Liu, Sruthi Sudhakar, Paarth Shah, Rares
  Ambrus, and Carl Vondrick.
\newblock Video generators are robot policies.
\newblock \emph{arXiv preprint arXiv:2508.00795}, 2025.

\bibitem[Liang et~al.(2024)Liang, Yu, Luo, Iyer, Dong, Zhou, Ghosh, Lewis, Yih,
  Zettlemoyer, et~al.]{liang2024mixture}
Weixin Liang, Lili Yu, Liang Luo, Srinivasan Iyer, Ning Dong, Chunting Zhou,
  Gargi Ghosh, Mike Lewis, Wen-tau Yih, Luke Zettlemoyer, et~al.
\newblock Mixture-of-transformers: A sparse and scalable architecture for
  multi-modal foundation models.
\newblock \emph{arXiv preprint arXiv:2411.04996}, 2024.

\bibitem[Lin et~al.(2026)Lin, He, Bao, Zhao, Li, Wang, Wang, Chi, and
  Zhang]{lin2026jepawam}
Yihan Lin, Jiawei He, Shifeng Bao, Chen Zhao, Yang Li, Xiaobo Wang, Yan Wang,
  Cheng Chi, and Jing Zhang.
\newblock {JEPA-WAM}: Learning vision-language-action policies with
  joint-embedding world modeling.
\newblock \emph{arXiv preprint arXiv:2608.09381}, 2026.
\newblock URL \url{https://arxiv.org/abs/2608.09381}.

\bibitem[Liu et~al.(2023)Liu, Zhu, Gao, Feng, Liu, Zhu, and
  Stone]{liu2023libero}
Bo~Liu, Yifeng Zhu, Chongkai Gao, Yihao Feng, Qiang Liu, Yuke Zhu, and Peter
  Stone.
\newblock {LIBERO}: Benchmarking knowledge transfer for lifelong robot
  learning.
\newblock \emph{Advances in Neural Information Processing Systems},
  36:\penalty0 44776--44791, 2023.

\bibitem[Liu et~al.(2025)Liu, Wu, Li, Tan, Chen, Wang, Xu, Su, and
  Zhu]{liu2025rdt}
Songming Liu, Lingxuan Wu, Bangguo Li, Hengkai Tan, Huayu Chen, Zhengyi Wang,
  Ke~Xu, Hang Su, and Jun Zhu.
\newblock {RDT-1B}: A diffusion foundation model for bimanual manipulation.
\newblock In \emph{International Conference on Learning Representations}, 2025.
\newblock URL \url{https://openreview.net/forum?id=yAzN4tz7oI}.

\bibitem[Ma et~al.(2026)Ma, Zheng, Wang, Jiang, Cui, Liang, and
  Yang]{ma2026dit4dit}
Teli Ma, Jia Zheng, Zifan Wang, Chunli Jiang, Andy Cui, Junwei Liang, and Shuo
  Yang.
\newblock {DiT4DiT}: Jointly modeling video dynamics and actions for
  generalizable robot control.
\newblock \emph{arXiv preprint arXiv:2603.10448}, 2026.

\bibitem[Mur-Labadia et~al.(2026)Mur-Labadia, Muckley, Bar, Assran, Sinha,
  Rabbat, LeCun, Ballas, and Bardes]{murlabadia2026vjepa21}
Lorenzo Mur-Labadia, Matthew Muckley, Amir Bar, Mido Assran, Koustuv Sinha,
  Mike Rabbat, Yann LeCun, Nicolas Ballas, and Adrien Bardes.
\newblock {V-JEPA 2.1}: Unlocking dense features in video self-supervised
  learning.
\newblock \emph{arXiv preprint arXiv:2603.14482}, 2026.

\bibitem[{NVIDIA} et~al.(2025)]{bjorck2025gr00t}
{NVIDIA} et~al.
\newblock {GR00T N1}: An open foundation model for generalist humanoid robots.
\newblock \emph{arXiv preprint arXiv:2503.14734}, 2025.

\bibitem[{Octo Model Team} et~al.(2024){Octo Model Team}, Ghosh, Walke,
  Pertsch, Black, Mees, Dasari, Hejna, Kreiman, Xu, Luo, Tan, Chen, Sanketi,
  Vuong, Xiao, Sadigh, Finn, and Levine]{octoteam2024octo}
{Octo Model Team}, Dibya Ghosh, Homer Walke, Karl Pertsch, Kevin Black, Oier
  Mees, Sudeep Dasari, Joey Hejna, Tobias Kreiman, Charles Xu, Jianlan Luo,
  You~Liang Tan, Lawrence~Yunliang Chen, Pannag Sanketi, Quan Vuong, Ted Xiao,
  Dorsa Sadigh, Chelsea Finn, and Sergey Levine.
\newblock Octo: An open-source generalist robot policy.
\newblock \emph{arXiv preprint arXiv:2405.12213}, 2024.
\newblock URL \url{https://arxiv.org/abs/2405.12213}.

\bibitem[Pai et~al.(2025)Pai, Achenbach, Montesinos, Forrai, Mees, and
  Nava]{pai2025mimic}
Jonas Pai, Liam Achenbach, Victoriano Montesinos, Benedek Forrai, Oier Mees,
  and Elvis Nava.
\newblock mimic-video: Video-action models for generalizable robot control
  beyond vlas.
\newblock \emph{arXiv preprint arXiv:2512.15692}, 2025.

\bibitem[Pan et~al.(2026)Pan, Peng, Zhang, Yan, Dai, Huo, Wang, Qi, Wang,
  Cheng, and Xu]{pan2026worldtowrist}
Yuhao Pan, Haosong Peng, Zhengshen Zhang, Zhengyang Yan, Yalun Dai, Fushuo Huo,
  Chujie Wang, Tianyu Qi, Xiucheng Wang, Nan Cheng, and Wenchao Xu.
\newblock {World-to-Wrist}: Task-conditioned future wrist modeling for
  fine-grained robot manipulation.
\newblock \emph{arXiv preprint arXiv:2608.05369}, 2026.
\newblock URL \url{https://arxiv.org/abs/2608.05369}.

\bibitem[{Physical Intelligence} et~al.(2025)]{black2025pi0.5}
{Physical Intelligence} et~al.
\newblock $\pi_{0.5}$: A vision-language-action model with open-world
  generalization.
\newblock \emph{arXiv preprint arXiv:2504.16054}, 2025.

\bibitem[Qu et~al.(2025)Qu, Song, Chen, Yao, Ye, Gu, Wang, Ding, Zhao, Wang,
  and Li]{qu2025spatialvla}
Delin Qu, Haoming Song, Qizhi Chen, Yuanqi Yao, Xinyi Ye, Jiayuan Gu, Zhigang
  Wang, Yan Ding, Bin Zhao, Dong Wang, and Xuelong Li.
\newblock {SpatialVLA}: Exploring spatial representations for
  visual-language-action models.
\newblock In \emph{Proceedings of Robotics: Science and Systems}, Los Angeles,
  CA, USA, June 2025.
\newblock \doi{10.15607/RSS.2025.XXI.011}.

\bibitem[Shao \& Li(2026)Shao and Li]{aiflowEdge}
Jiawei Shao and Xuelong Li.
\newblock {AI Flow} at the network edge.
\newblock \emph{IEEE Network}, 40\penalty0 (1):\penalty0 330--336, 2026.
\newblock \doi{10.1109/MNET.2025.3541208}.

\bibitem[Song et~al.(2025)Song, Zhou, Zhao, Chen, Ding, Yan, Huang, Tang, Wang,
  and Li]{song2025reconvlareconstructivevisionlanguageactionmodel}
Wenxuan Song, Ziyang Zhou, Han Zhao, Jiayi Chen, Pengxiang Ding, Haodong Yan,
  Yuxin Huang, Feilong Tang, Donglin Wang, and Haoang Li.
\newblock Reconvla: Reconstructive vision-language-action model as effective
  robot perceiver, 2025.
\newblock URL \url{https://arxiv.org/abs/2508.10333}.

\bibitem[Sun et~al.(2026)Sun, Zhang, Qi, Ren, Liu, Zhu, Sun, Jin, and
  Chen]{sun2026vla}
Jingwen Sun, Wenyao Zhang, Zekun Qi, Shaojie Ren, Zezhi Liu, Hanxin Zhu,
  Guangzhong Sun, Xin Jin, and Zhibo Chen.
\newblock {VLA-JEPA}: Enhancing vision-language-action model with latent world
  model.
\newblock \emph{arXiv preprint arXiv:2602.10098}, 2026.

\bibitem[Wan et~al.(2025)Wan, Wang, Ai, Wen, Mao, Xie, Chen, Yu, Zhao, Yang,
  et~al.]{wan2025wan}
Team Wan, Ang Wang, Baole Ai, Bin Wen, Chaojie Mao, Chen-Wei Xie, Di~Chen,
  Feiwu Yu, Haiming Zhao, Jianxiao Yang, et~al.
\newblock Wan: Open and advanced large-scale video generative models.
\newblock \emph{arXiv preprint arXiv:2503.20314}, 2025.

\bibitem[Won et~al.(2025)Won, Lee, Jang, Kim, and
  Shin]{won2025dualstreamdiffusionworldmodelaugmented}
John Won, Kyungmin Lee, Huiwon Jang, Dongyoung Kim, and Jinwoo Shin.
\newblock Dual-stream diffusion for world-model augmented
  vision-language-action model, 2025.
\newblock URL \url{https://arxiv.org/abs/2510.27607}.

\bibitem[Wu et~al.(2023)Wu, Jing, Cheang, Chen, Xu, Li, Liu, Li, and
  Kong]{wu2023unleashing}
Hongtao Wu, Ya~Jing, Chilam Cheang, Guangzeng Chen, Jiafeng Xu, Xinghang Li,
  Minghuan Liu, Hang Li, and Tao Kong.
\newblock Unleashing large-scale video generative pre-training for visual robot
  manipulation, 2023.

\bibitem[Wu et~al.(2026)Wu, Lu, Wang, Yang, Liu, Wang, Zhu, Sun, Wang, Ma,
  et~al.]{wu2026pragmatic}
Wei Wu, Fan Lu, Yunnan Wang, Shuai Yang, Shi Liu, Fangjing Wang, Qian Zhu,
  He~Sun, Yong Wang, Shuailei Ma, et~al.
\newblock A pragmatic vla foundation model.
\newblock \emph{arXiv preprint arXiv:2601.18692}, 2026.

\bibitem[Yu et~al.(2026)Yu, Lin, Zhang, Zhang, Gu, Li, and Tan]{maskwam}
Hanyang Yu, Haitao Lin, Jingbo Zhang, Wenyao Zhang, Chenghao Gu, Heng Li, and
  Ping Tan.
\newblock {MaskWAM}: Unifying mask prompting and prediction for world-action
  models.
\newblock \emph{arXiv preprint arXiv:2606.13515}, 2026.

\bibitem[Yuan et~al.(2026{\natexlab{a}})Yuan, Zhao, Shi, Jiang, Guo, Liu, Liu,
  Sui, and Wang]{yuan2026dreamwam}
Shanglin Yuan, Weiheng Zhao, Xin Shi, Haoyi Jiang, Xianda Guo, Liu Liu, Wenyu
  Liu, Wei Sui, and Xinggang Wang.
\newblock {DreamWAM}: Beyond {RGB} future prediction for world action models.
\newblock \emph{arXiv preprint arXiv:2608.04996}, 2026{\natexlab{a}}.
\newblock URL \url{https://arxiv.org/abs/2608.04996}.

\bibitem[Yuan et~al.(2026{\natexlab{b}})Yuan, Dong, Liu, and
  Zhao]{yuan2026fast}
Tianyuan Yuan, Zibin Dong, Yicheng Liu, and Hang Zhao.
\newblock {Fast-WAM}: Do world action models need test-time future imagination?
\newblock \emph{arXiv preprint arXiv:2603.16666}, 2026{\natexlab{b}}.

\bibitem[Zhang et~al.(2025)Zhang, Liu, Qi, Wang, Yu, Zhang, Dong, He, Lu, Wang,
  Zhang, Yi, Zeng, and Jin]{dreamvla25}
Wenyao Zhang, Hongsi Liu, Zekun Qi, Yunnan Wang, Xinqiang Yu, Jiazhao Zhang,
  Runpei Dong, Jiawei He, Fan Lu, He~Wang, Zhizheng Zhang, Li~Yi, Wenjun Zeng,
  and Xin Jin.
\newblock Dreamvla: A vision-language-action model dreamed with comprehensive
  world knowledge.
\newblock \emph{CoRR}, abs/2507.04447, 2025.
\newblock \doi{10.48550/ARXIV.2507.04447}.
\newblock URL \url{https://doi.org/10.48550/arXiv.2507.04447}.

\bibitem[Zhang et~al.(2026)Zhang, Zhang, Qi, Zeng, Jin, and
  Zhang]{zhang2026disentangled}
Wenyao Zhang, Bozhou Zhang, Zekun Qi, Wenjun Zeng, Xin Jin, and Li~Zhang.
\newblock Disentangled robot learning via separate forward and inverse dynamics
  pretraining.
\newblock \emph{arXiv preprint arXiv:2604.16391}, 2026.

\bibitem[Zheng et~al.(2025)Zheng, Li, Wang, Liu, Kang, Feng, Zheng, Zou, Chen,
  Zeng, Zhang, Pang, Liu, Wang, and Zhan]{zheng2025xvla}
Jinliang Zheng, Jianxiong Li, Zhihao Wang, Dongxiu Liu, Xirui Kang, Yuchun
  Feng, Yinan Zheng, Jiayin Zou, Yilun Chen, Jia Zeng, Ya-Qin Zhang, Jiangmiao
  Pang, Jingjing Liu, Tai Wang, and Xianyuan Zhan.
\newblock {X-VLA}: Soft-prompted transformer as scalable cross-embodiment
  vision-language-action model, 2025.
\newblock URL \url{https://arxiv.org/abs/2510.10274}.
\newblock arXiv preprint arXiv:2510.10274.

\bibitem[Zhou et~al.(2024)Zhou, Du, Chen, Li, Yeung, and
  Gan]{zhou2024robodreamer}
Siyuan Zhou, Yilun Du, Jiaben Chen, Yandong Li, Dit-Yan Yeung, and Chuang Gan.
\newblock {RoboDreamer}: Learning compositional world models for robot
  imagination.
\newblock In \emph{Proceedings of the 41st International Conference on Machine
  Learning}, volume 235 of \emph{Proceedings of Machine Learning Research},
  pp.\  61885--61896. PMLR, 2024.
\newblock URL \url{https://proceedings.mlr.press/v235/zhou24f.html}.

\bibitem[Zhu et~al.(2025)Zhu, Yu, Feng, Burchfiel, Shah, and
  Gupta]{zhu2025unifiedworldmodelscoupling}
Chuning Zhu, Raymond Yu, Siyuan Feng, Benjamin Burchfiel, Paarth Shah, and
  Abhishek Gupta.
\newblock Unified world models: Coupling video and action diffusion for
  pretraining on large robotic datasets.
\newblock In \emph{Proceedings of Robotics: Science and Systems}, Los Angeles,
  CA, USA, June 2025.
\newblock \doi{10.15607/RSS.2025.XXI.015}.

\end{thebibliography}
\bibliographystyle{iclr2027_conference}

\end{document}